\documentclass[letterpaper, 10 pt, conference]{packages/ieeeconf}

\IEEEoverridecommandlockouts
\usepackage[skip=2pt,font=footnotesize]{caption} 
\usepackage{multirow}
\usepackage{graphics} 
\usepackage{epsfig} 
\usepackage{times} 
\usepackage{amssymb}  
\usepackage{mathtools}
\usepackage{packages/commath}

\usepackage{amsthm}
\usepackage{graphicx}
\usepackage{subfloat}
\usepackage{subcaption}

\usepackage[colorlinks=true]{hyperref}
 \usepackage[font=small]{caption}

\usepackage{makecell}
\usepackage{tikz}

\usepackage{array}
\newcolumntype{M}[1]{>{\centering\arraybackslash}m{#1}}

\usepackage{colortbl}
\usepackage{tabulary}

\usepackage{amsmath,amsfonts} 

\usepackage{amsmath,amssymb,amsfonts}
\usepackage{graphicx}
\DeclareGraphicsExtensions{.eps,.pdf,.png,.jpg,.JPG,.gif}
\usepackage{subfloat}
\usepackage{bm}
\usepackage[colorlinks=true]{hyperref}
\usepackage{url}
\usepackage{xcolor}

\newcolumntype{M}[1]{>{\centering\arraybackslash}m{#1}}

\usepackage{array}
\usepackage{booktabs}
\newcommand{\PreserveBackslash}[1]{\let\temp=\\#1\let\\=\temp}
\newcolumntype{C}[1]{>{\PreserveBackslash\centering}m{#1}}
\newcolumntype{R}[1]{>{\PreserveBackslash\raggedleft}p{#1}}
\newcolumntype{L}[1]{>{\PreserveBackslash\raggedright}p{#1}}

\usepackage[backend=bibtex,natbib=true,style=numeric-comp,sorting=none,firstinits=true,maxbibnames=99,url=false,doi=false]{biblatex}

\newcommand\scalemath[2]{\scalebox{#1}{\mbox{\ensuremath{\displaystyle #2}}}}

\usepackage{xcolor}

\usepackage{packages/algorithm}
\usepackage{packages/algorithmic}\usepackage{array} 
\usepackage{adjustbox}

\newcommand{\bx}{\mathbf{x}}

\newcommand{\bp}{\mathbf{p}}

\newcommand{\bc}{\mathbf{c}}

\newcommand{\bI}{\mathbf{I}}
\newcommand{\bR}{\mathbf{R}}
\newcommand{\bD}{\mathbf{D}}

\newcommand{\be}{\mathbf{e}}
\newcommand{\bS}{\mathbf{S}}
\newcommand{\bM}{\mathbf{M}}

\usepackage{mathtools}

\DeclarePairedDelimiter\floor{\lfloor}{\rfloor}
\allowdisplaybreaks 
\renewcommand{\baselinestretch}{0.96}
\newcommand{\bl}[1]{\color{blue}{\bf{#1}}}
\newcommand{\gr}[1]{\color{green!75!magenta}{\bf{#1}}}

\title{\Large \bf
CodeVIO: Visual-Inertial Odometry with Learned Optimizable Dense Depth
}
\author{Xingxing Zuo$^{1,2,\dagger}$, Nathaniel Merrill$^{3,\dagger}$, Wei Li$^{4}$, Yong Liu$^2$,  Marc Pollefeys$^{1,5}$, Guoquan Huang$^3$
    \thanks{This work was partially supported by the Department of Computer Science at ETHz, the National Natural Science Foundation of China under Grant 61836015, the  University  of Delaware (UD) College of Engineering, the ARL (W911NF-19-2-0226),  and  Google  ARCore. Xingxing Zuo was partially supported by the Chinese Scholarship Committee. (Yong Liu is the corresponding author, email: yongliu@iipc.zju.edu.cn)}
\thanks{$^1$ Department of Computer Science,  ETH Z\"{u}rich.}%
\thanks{$^2$ Institute of Cyber-System and Control, Zhejiang University.}%
\thanks{$^3$ Robot Perception and Navigation Group, University of Delaware.}%
\thanks{$^4$ Inceptio Technology, Shanghi, China.}%
\thanks{$^5$ Microsoft Mixed Reality and Artificial Intelligence Lab, Z\"{u}rich.}%
\thanks{$^\dagger$ These authors contributed equally to this work.}%
}

\begin{document}
	
\maketitle

\begin{abstract}
	In this work, we present a lightweight, tightly-coupled deep depth network and visual-inertial odometry (VIO) system, which can provide accurate state estimates and dense depth maps of the immediate surroundings.
	Leveraging the proposed lightweight Conditional Variational Autoencoder (CVAE) for depth inference and encoding, we provide the network with previously marginalized sparse features from VIO to increase the accuracy of initial depth prediction and generalization capability. 
	The compact representation of dense depth, termed depth code, can be 
	updated jointly with navigation states 
	in a sliding window estimator in order to provide the dense 
	local scene geometry.
	We additionally propose a novel method to obtain the CVAE's Jacobian
	which is shown to be more than an order of magnitude faster than previous works,
	and we additionally leverage First-Estimate Jacobian (FEJ) to avoid
	recalculation.
	As opposed to previous works that rely on completely dense residuals,
	we propose to only provide sparse measurements to update the depth code
	and show through careful experimentation that our choice of sparse measurements and FEJs can still significantly improve the estimated depth maps.
	Our full system also exhibits state-of-the-art pose estimation accuracy, and we show that it can run in real-time with single-thread execution while utilizing GPU acceleration only for the network and code Jacobian.
\end{abstract}

\section{Introduction}
Accurate pose estimation and dense depth estimation are essential to a wide range of robotic applications such as obstacle avoidance and path planning.
Recently, utilizing Conditional Variational Autoencoder (CVAE)
networks as a differentiable manifold 
to compress dense depth maps into a far lower-dimensional subspace, has 
shown promising results \cite{bloesch2018codeslam} and even shown real-time 
performance when leveraging a desktop graphics card and incremental 
smoothing \cite{czarnowski2020deepfactors}.
However, these methods only utilize camera information, and rely solely on 
the depth network to recover scale information from learned priors.
Additionally, even the real-time method \cite{czarnowski2020deepfactors}
still relies heavily on GPU acceleration to perform dense warping 
to extract whole-image photometric residuals, and to our knowledge,
no work thus far has presented a method to efficiently 
calculate the network Jacobian.

\begin{figure} [!t]
	\centering
	\includegraphics[width=\columnwidth, height=0.5\columnwidth]{%
		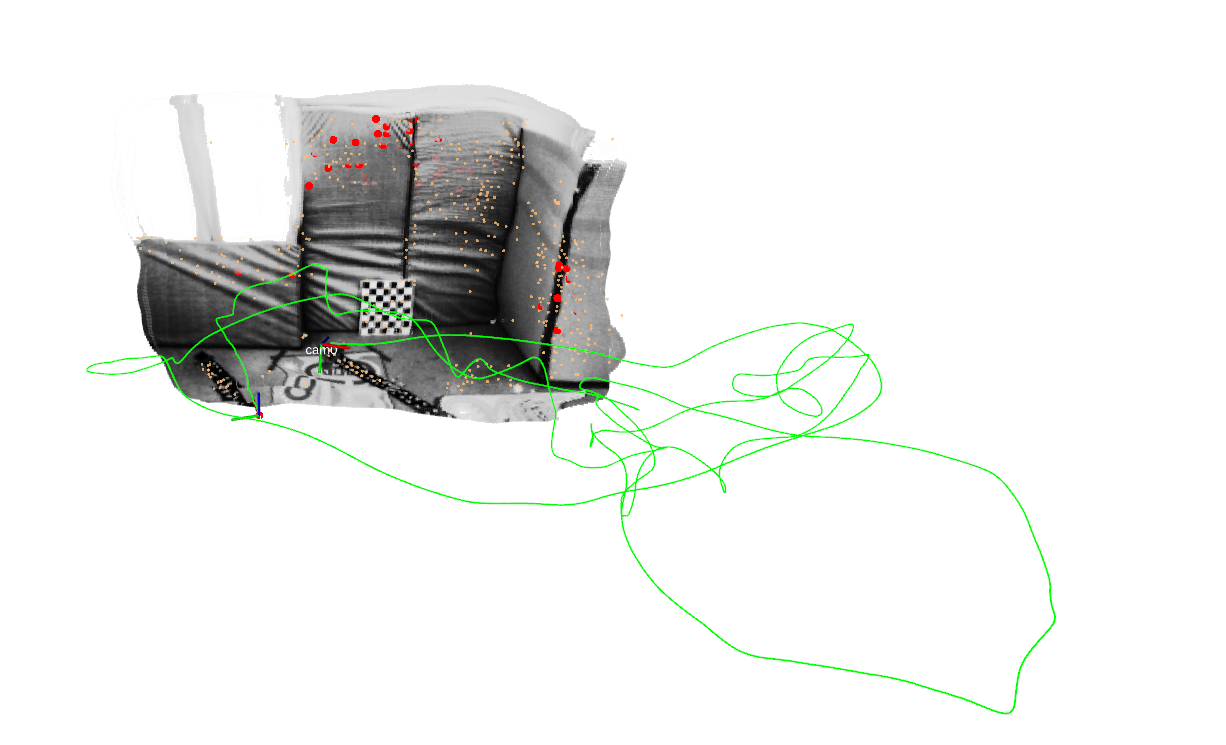}
	
	\includegraphics[width=\columnwidth]{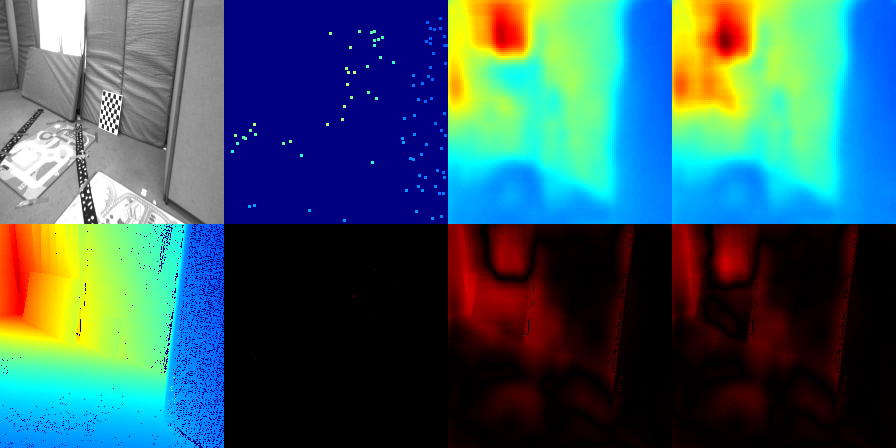}
	\caption{
		{\bf Top:} An example 3-view reconstruction on the EuRoC dataset.
		{\bf Bottom:} The top row from left to right shows
		the input image and sparse depths, depth map before and after sparse updates.
		On the bottom, the projected LiDAR map serves as a ground truth comparison, producing the heatmaps of the depth estimation error -- which is greatly reduced after updates of depth code.
	}
	\label{fig:overview}
	\vspace*{-2em}
\end{figure}

To address some of these shortcomings, we take a completely different
approach, and leverage a lightweight network and IMU-driven 
EKF-based estimator to gain some easy speedups and provide a metric
scale to the whole system.
Furthermore, we propose a new method to recover the approximate 
VAE decoder Jacobian with near-perfect accuracy and over an order of magnitude speedup over the traditional chainrule method provided by deep learning libraries, and show that only
providing sparse measurements to the VAE can still 
improve performance considerably and in an efficient manner.
Our main contributions are:
\begin{itemize}
	\item We develop an efficient VIO system that jointly estimates
	a scaled local dense reconstruction along with sparse 
	features and IMU poses in a tightly-coupled manner over
	a sliding window in real time
	using single thread
	CPU execution and only utilizing the GPU to accelerate
	the network inference and VAE Jacobian.
	\item A lightweight CVAE model is proposed, which leverages
	depthwise-separable convolutions and additive skip connections
	to enable highly-efficient inference while retaining a high level 
	of accuracy. Through CVAE, a dense depth map is encoded into a low-dimensional depth code, which can be jointly updated with navigation states.
	\item We present a new method to recover the approximate VAE decoder
	Jacobian, which  has near-perfect accuracy and
	achieves over an order of magnitude speedup over the 
	chainrule method previously adopted.
	Also, due to our use of First-Estimate Jacobians (FEJ),
	the Jacobian never has to be recalculated for any image.
	\item Due to our fusion of IMU information, 
	we propose to feed marginalized sparse features from recent
	frames along with the images to the network to increase
	the accuracy of the initial depth map and thus the 
	zero code prior.
	Furthermore, we only use sparse updates on the codes
	that are estimated in order to further reduce the required
	computation and necessity for GPU acceleration.
\end{itemize}

\section{Related Work}

Mapping with the aid of IMU is appealing. 
A dense mapping method deployed on aerial robots is proposed in~\cite{yang2017real}. Poses are estimated in a factor graph optimization-based VIO. Then the scale determined poses from VIO are passed into a motion stereo based depth estimation module. 
In this way, the scale information from IMU is just implicitly incorporated into the dense depth through motion stereo.
Kimera~\cite{rosinol2020kimera} is another VI-SLAM system with the capability of mesh reconstruction, which supports both monocular and stereo camera.
Along with pose estimation, a 3D mesher module for dense reconstruction is enabled by sparse points aided 2D Delaunay triangulation.

Without additional sensors, neural networks can also recover monocular depth with pseudo scale, which is learned in a data-driven fashion, such as~\cite{lasinger2019towards, wofk2019fastdepth}.
In FastDepth~\cite{wofk2019fastdepth}, an efficient and lightweight encoder-decoder network architecture for depth prediction is proposed. It is fast and applicable to embedded systems for real-time depth prediction. 
However, the accuracy and reliability of monocular depth prediction can not be guaranteed in practical application. To address this issue, some works resort to scale determined sparse depth. For example,
Ma and Karaman~\cite{ma2018sparse} propose a deep CNN 
for dense depth prediction from monocular image and sparse depth,
%
while Tang et al.~\cite{tang2019learning} propose a learning guided convolutional network for depth completion. 
Similar to our work, VIO sparse depths are utilized for depth completion by Wong et al.~\cite{wong2020unsupervised, wong2021learning}.

Due to the significant progress in learning-based depth prediction, SLAM with learned dense depth are also explored in many existing works~\cite{tateno2017cnn, yang2018deep, zhou2018deeptam, lee2020real}.
CNN-SLAM~\cite{tateno2017cnn} leverages single image depth prediction and perform dense alignment for a dense 3D reconstruction in a full SLAM system. 
In~\cite{yang2018deep}, deep depth predictions are incorporated into a directed method based monocular visual optometry as direct virtual stereo measurements for dense mapping.
However, in all the method mentioned above, the dense depth are computationally intensive to process due to its high dimensionality. 
To address this issue, a milestone method, CodeSLAM~\cite{bloesch2018codeslam}, is proposed.
CodeSLAM introduces the concept of learning a compact optimizable representation of dense depth through conditional variational auto-encoder (CVAE). Dense depth is encoded in a compact code and can be dexterously incorporated for solving a dense structure-from-motion problem.
CodeSLAM is computationally intensive, and thus its follow-up work DeepFactors~\cite{czarnowski2020deepfactors} is developed with a capability of real-time SLAM in a factor graph-based optimization framework. 
However, a big portion of computation in the system, including image warping
and full-image tracking, is conducted on the GPU, 
and the network Jacobian calculation is still a bottleneck.
CodeSLAM and DeepFactors are the two closest work to our CodeVIO. However, both can only perform up-to-scale dense reconstruction, since only RGB images are utilized in the inference stage. 
In contrast, our proposed CodeVIO offers metric dense depth and shows great generalization capability.

\section{Fast Depth Inference and Encoding} \label{sec:fastdepth}

\begin{figure} [!t]
	\centering
	\includegraphics[width=\columnwidth]{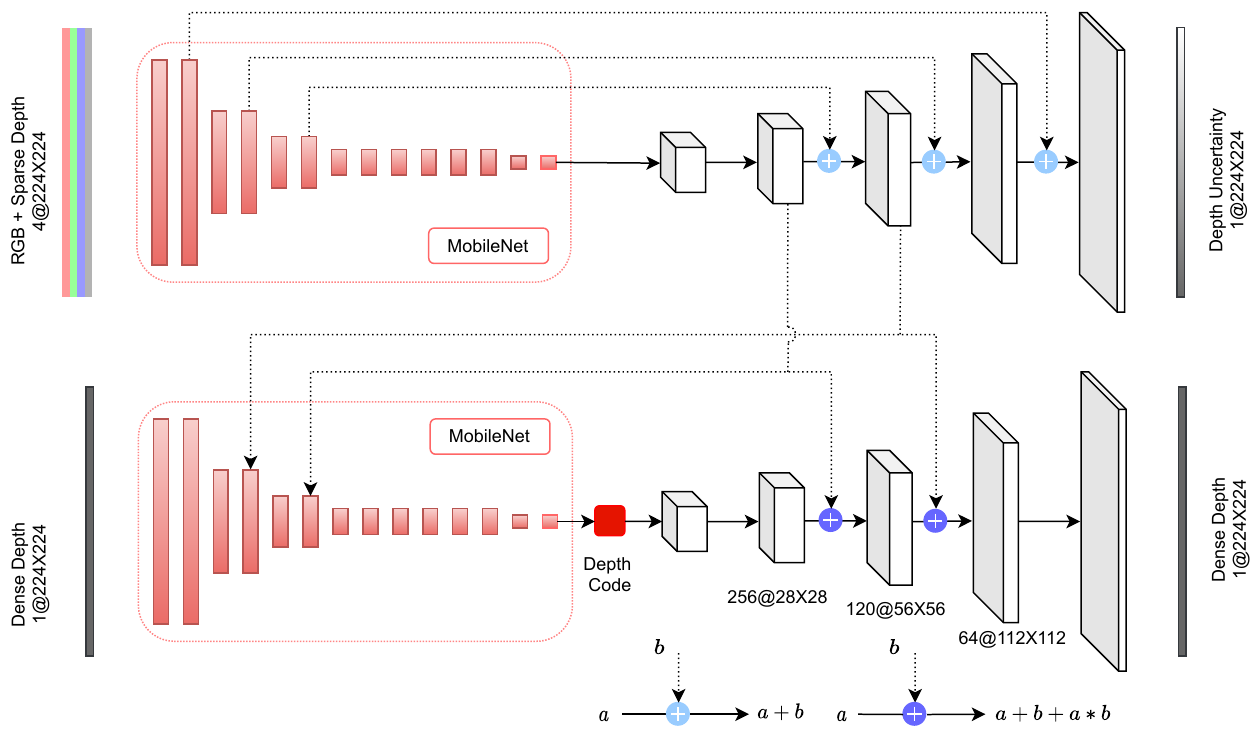}
	\caption{ The architecture of the conditional variational auto-encoder (CVAE) consisting of two streams: a UNet to extract features from 
		image and sparse depth (top stream), 
		and a variational auto-encoder (VAE) conditioned on the UNet features for depth encoding and decoding (bottom stream).
		Dimensions are denoted as channels $@$ height $\times$ width. 
	}
	\label{fig:network}
	\vspace*{-2em}
\end{figure}


Inspired by CodeSLAM~\cite{bloesch2018codeslam}, the proposed network architecture contains two streams of information (see Fig.~\ref{fig:network}):
(i) a modified FastDepth~\cite{wofk2019fastdepth} with the pruned architecture,
where the input of RGB or grayscale image concatenated with a sparse depth map from VIO is used to obtain deep feature map for our second stream as well as estimate a depth inference uncertainty map;
and (ii) a variational auto-encoder (VAE),
which is conditioned on the learned deep features from the first stream, and is trained to encode dense depth maps. 
During training and validation, we sample sparse depth points from the groundtruth dense depth maps to allow training on a general RGBD dataset.
To imitate the sparse depths from VIO, for training, we detect FAST corner features~\cite{rosten2006machine} from the images and randomly sample $50 - 200$ of them. 
To mitigate the sensitivity to the noisy depth from VIO,
we inject Gaussian noises with a standard deviation of $0.1m$ to the sampled sparse depth in the training phase. 
This sparse depth sampling strategy mostly avoids overfitting to idealistic uniformly-distributed true depth values with
fixed sparsity as in existing literature~\cite{ma2018sparse}.

To ensure low latency, all the convolutional layers are depthwise separable~\cite{howard2017mobilenets}, instead of standard convolutional layers. 
An efficient MobileNet~\cite{howard2017mobilenets} with depthwise decomposition serves as the encoders in both the UNet feature extraction stream and the auto-encoder stream. 
The upsample layers consist of depthwise separable convolutions with a kernel size of 5 and nearest neighbour based interpolation. 
The output of the upsample layers has decreased channels compared to the input channels.
In the VAE stream, dense depth is encoded into a low dimensional depth code at the bottleneck of VAE in the training phase. 
Then dense depth can be decoded from the depth code. 
In the inference phase, as dense depth is unknown,
the VAE encoder is ignored. The dense depth map can be predicted from the zero code conditioned on the deep feature maps from the UNet. 
In initial experimentation, we toyed with leveraging the linear decoder
idea of~\cite{bloesch2018codeslam}, however, due to the
proposed novel method to efficiently obtain the Jacobian as
well as our use of FEJ~\cite{Huang2008ICRA,Huang2009ERB,Huang2010IJRR},
we actually do not explicitly 
use any property of the linear decoder, and simply leave it in our
network as it did not significantly hurt the accuracy.

To  lower computations, 
all the skip connections are via addition rather than concatenation as in CodeSLAM~\cite{bloesch2018codeslam}.
The skip connections provide low-level gradient and scale information 
from the UNet encoder for its decoder to preserve details
from the input image and sparse depth.
Similar to \cite{bloesch2018codeslam}, to preserve the influence of UNet feature maps on the VAE's code Jacobian, 
we additionally add the element-wise multiplication term to the additive skip connection,
thus compressing the concatenated skip connections of CodeSLAM by a factor of three while preserving most of the information. 
Similar to \cite{wofk2019fastdepth}, we found that the additive skip
connections speed up the network significantly with little
affect on accuracy.
We also found that two skip connections with dimensions of $256@28\times28$ (channels $@$ height $\times$ width) and  $120@56\times56$ are enough for the VAE stream to merge the information from the UNet stream.

The training loss in our network is the summation of the depth recovery loss and the VAE's KL divergence loss~\cite{kingma2013auto}.
The KL divergence loss closes the gap of the depth code distribution to standard normal distribution. 
We predict the logarithm of the inverse depth values ($\hat{\mathbf D}_{loginv}$) from the VAE rather than the raw depth values or the proximity proposed in \cite{bloesch2018codeslam}, 
and the UNet predicts the logarithm of depth uncertainty value 
%
($\hat{\mathbf B}_{log}$). 
The depth recovery loss is given by:
\begin{align}
	\scalemath{.9}{
		L_{rec} = \frac{1}{|\Omega|} \sum_{x \in \Omega}
		\frac{|\hat{D}_{loginv}(x) - \log^{-1}(D(x))|
		}{\exp({\hat{B}_{log}(x)})}
		+ \hat{B}_{log}(x)
	}
\end{align}
where $x \in \Omega$ denotes every pixel in the output feature maps.
The loss is averaged for each batch during training, and the 
depth values are thresholded accordingly to prevent an infinite $\log$.
To stabilize  training, we leverage a KL annealing strategy like~\cite{bowman2015generating, zhi2019scenecode},
where the weight on KL divergence loss term is steadily increased from zero after three epochs.

\section{Probabilistic VIO with Optimizable Depth} \label{sec:vio}


Leveraging 
OpenVINS~\cite{Geneva2020ICRA},
we perform monocular VIO {\em and} local dense mapping with 
an extension of the Multi-State Constraint Kalman Filter (MSCKF) \cite{mourikis2007multi}. 
The state vector consists of the IMU navigation state $\mathbf{x}_{I}$, a set of $m$ historical IMU clone poses $\mathbf{x}_{clone}$, 
$\ell$ ``long-lived'' sparse visual feature points $\bx_{f}$, the calibration state $\mathbf{x}_{calib}$ (spatial-temporal, camera intrinsics), as well
as a set of $n = \floor{m / p}$ CVAE codes $\bx_{code}$, where $n \leq m$ as we keyframe the codes by only initializing them for every $p$th image. 
\begin{align}
	\label{eq:state vector}
	\mathbf{x} & = 
	\begin{bmatrix}
		\mathbf{x}^{\top}_{I} & \mathbf{x}^{\top}_{clone} &
		\bx_{f}^{\top} & \mathbf{x}^{\top}_{calib} &  \bx_{code}^{\top}
	\end{bmatrix}^{\top} \\
	\mathbf{x}_{I} & = 
	\begin{bmatrix}
		{}^{I_k}_G\bar{q}^{\top} & \mathbf{b}^{\top}_{g} & {}^G\mathbf{v}^{\top}_{I_k} & \mathbf{b}^{\top}_{a} & {}^G\mathbf{p}^{\top}_{I_k}
	\end{bmatrix}^{\top}  \\
	\mathbf{x}_{clone} &=
	\begin{bmatrix}
		{}^{I_{k-1}}_G\bar{q}^{\top}  &   {}^G\mathbf{p}_{I_{k-1}}^{\top} &\!   \cdots  \!&  {}^{I_{{k-m}}}_G\bar{q}^{\top} &  {}^G\mathbf{p}_{I_{{k-m}}}^{\top} 
	\end{bmatrix}^\top \\
	\label{eq:state calib}
	\mathbf{x}_{calib} & = 
	\begin{bmatrix}
		{}^C_I\bar{q}^{\top} &  {}^C\mathbf{p}^{\top}_{I} & t_{d}
		& \bm{\zeta}^\top
	\end{bmatrix}^{\top} \\
	\bx_{f} &=
	\begin{bmatrix}
		{}^A\bp_{f_0}^{\top}  &  {}^A\bp_{f_1}^{\top}  &\!   \cdots  \!&  {}^A\bp_{f_{\ell-1}}^{\top} 
	\end{bmatrix}^\top  \\
	\bx_{code} &=
	\begin{bmatrix}
		\bc_{k}^{\top}  & \bc_{k-p}^{\top}  &\! \cdots  \!&  \bc_{{k-n p}}^{\top} 
	\end{bmatrix}^\top
\end{align}
In the above expressions, besides the standard VIO notations that can be found in \cite{Geneva2020ICRA}, we here focus on the new state, i.e., CVAE code $\bx_{code}$.
%
The depth codes  are extremely low-dimensional compared to the depth maps they encode (32 vs $224^2=50176$ in our case) and can be efficiently updated with sparse measurements in the sliding-window, which is considerably different than the dense measurements used by similar works. 
While the codes are low dimensional, including one for every image still adds considerable
computation to the EKF update, and is highly redundant -- as the depth maps will overlap 
more than necessary.
To this end, 
we use a constant interval keyframing strategy in order to simplify the logic of ensuring a code is always in the sliding window.

\subsection{Optimizable Dense Depth and Inertial Navigation}

We perform state estimation and dense local mapping in the MSCKF framework, 
leveraging FEJ
to improve consistency as well as negate the need to ever re-calculate a code's Jacobian.
The IMU propagation phase is the standard forward 
integration in MSCKF. 
The IMU states $\bx_{I}$ and their corresponding 
covariances are propagated from the last image time 
instant in IMU time axis to the current time.
Then, stochastic cloning of the propagated IMU pose is performed for state augmention.
%
%
If the image frame is selected as a keyframe, we will also augment the state vector with the depth code of this image.
The depth code is initialized to be a zero vector
and assigned a covariance of $\sigma^2_{c}\mathbf{I}$,
which coincides with the assumption that the depth code in the VAE is under the standard normal distribution.
In practice, we use $\sigma_c > 1$ to account 
for the fact that the VAE does not 
\textit{perfectly} follow the standard normal distribution.
%
All the states, including the depth codes, are updated with measurements in the update phase.

By encoding the dense depth of keyframes with the low dimensional depth code via VAE, the dense depth is a function of the raw image $\bM$, sparse depth $\bS$, and the depth code estimate $\hat{\bc}$.
Thus we write 
\begin{align}
	\hat{\bD} ~ 
	\dot{=} ~ f(\hat{\bc}) ~
	\dot{=} ~ f(\bM, \bS, \hat{\bc}) 
\end{align}
in order to simplify notation.
Here $f$ is simply the VAE decoder function.
The depth map is a nonlinear function of $\bM$ and $\bS$
due to nonlinear activations in the UNet.
It is also important to note that, although a linear decoder is employed in the VAE, the final operation to obtain the network output in depth form $\hat{D}[u,v] = \exp(-\hat{D}_{loginv}[u,v])$
makes the depth map a nonlinear function of the code vector as well.

\subsubsection{Reprojection Error and Sparse Geometric Update}
\label{sec:reprojection error}
For a triangulated 3D sparse visual feature  ${}^{A}\mathbf{p}_{f_i}$ observed in image $\bM_j$ (at camera frame $\{C_j\}$) with observation $\mathbf{z}^{j}_{rep,f_i} = [u^j_i,v^j_i]^\top$, the reprojection residual is~\cite{hartley2003multiple}:
\begin{align}
	\mathbf{r}^{rep}_{i,j} &= 
	\mathbf{z}^{j}_{rep,f_i} \!\!-\!\! 
	h(\hat{\mathbf{x}}_j, 
	{}^{A}\hat{\mathbf{p}}_{f_i}) 
	\approx 
	\mathbf{H}_{{x}_j} \tilde{\mathbf{x}}_j \!+\!
	\mathbf{H}_{f_i} {}^{A}\tilde{\mathbf{p}}_{f_i} 
	\!+\! \mathbf{n}_r
	\label{eq:reprojection error}
\end{align}
where 
$\scalemath{0.9}{\mathbf{x}_j = \begin{bmatrix} 
		{}^{I_{j}}_G\bar{q}^{\top}  &       {}^G\mathbf{p}_{I_{j}}^{\top} &
		{}^C_I\bar{q}^{\top} &
		{}^C\mathbf{p}^{\top}_{I} &
		\bm{\zeta}
	\end{bmatrix}^\top} $
includes the cloned pose corresponding to $\{C_j\}$ as well as the necessary 
the camera intrinsics $\bm{\zeta}$ and the IMU-to-camera 
extrinsic parameters to project ${}^A\mathbf{p}_{f_i}$ onto the image plane.
Additionally, if the anchor frame $\{A\}$ is not the static global frame,
then $\mathbf{x}_j$ also contains the camera clone pose
corresponding to $\{A\}$.
$\mathbf{H}_{{x}_j}$ and $\mathbf{H}_{f_i}$ are the
Jacobians of the reprojection error residual
with respect to $\bx_j$ and ${}^{A}\mathbf{p}_{f_i}$.
$\mathbf{n}_r$ denotes the feature tracking noise 
under a Gaussian distribution with 
isotropic covariance matrix 
$\bR = \sigma^2_{im}\bI_2$.

If  $\{C_j\}$ is also a keyframe with a corresponding depth map
${}^{C_j}\bD$, we can formulate a sparse geometric
constraint for the observed visual feature
${}^{A}\mathbf{p}_{f_i}$,
by now considering $\mathbf{z}^j_{rep,f_i}$
as the noise-free $[u ~ v]^\top$ 
(dropping the subscripts to simplify notation),
and utilizing the observation of 
the depth value 
${}^{C_j}\hat{D}[u,v]$
at the nearest pixel location of the depth map,
which is of course noisy due to the network weights,
we can write the sparse geometric residual as
\begin{align}
	r^{geo}_{i,j} &= {}^{C_j}{D}[u,v] -   
	g(\hat{\mathbf{x}}_j,{}^{A}\hat{\mathbf{p}}_{f_i}) \notag \\
	&= 
	{}^{C_j}\hat{D}[u,v]
	- \be^\top_3 \left( {}^{C_j}_{A}\hat{\bR} {}^{A}\hat{\mathbf{p}}_{f_i} 
	+ {}^{C_j}\hat{\bp}_{A}  \right) 
	+ n^j_{d_i} \notag \\
	&\approx 
	\mathbf{G}_{{c}_j} \tilde{\mathbf{c}}_{j} +
	\mathbf{G}_{{x}_j} \tilde{\mathbf{x}}_j + \mathbf{G}_{f_i} {}^{A}\tilde{\mathbf{p}}_{f_i} + n^j_{d_i}
	\label{eq:geometry error}
\end{align}
where $\be_3 = [0 ~ 0 ~ 1]^\top$,
and ${}^{C_j}_{A}\hat{\bR}, {}^{C_j}\hat{\bp}_{A}$
are functions of $\hat{\mathbf{x}}_j$.
%
Moreover, $\mathbf{G}_{{c}_j}$, $\mathbf{G}_{x_j}$ and $\mathbf{G}_{f_j}$ are the Jacobians of depth geometry measurement with respect to the error states of ${\mathbf{c}}_{j}$,  ${\mathbf{x}}_j$, and ${}^{A}{\mathbf{p}}_{f_i}$,
respectively. 
In this case,
$\mathbf{G}_{c_j}$ is a single row of the 
Jacobian of predicted depth with respect to its code as described in Sec.~\ref{sec:network_Jacobian}, which is
evaluated at the corresponding row for $[u ~ v]^\top$.
The noise of the depth geometry measurement is modeled by
$n^j_{d_i} \sim \mathcal(0,\sigma^2_{i,j})$, where
$\sigma_{i,j} ~ \dot{=} ~ 
\exp({}^{C_j}\hat{B}_{log}[u^j_i,v^j_i])$
and thus the measurement covariance can be directly fetched
from the predicted depth uncertainty. In contrast to reprojection noise, the depth noise generally
varies for different observations of the same feature 
( i.e. $\sigma^2_{i,a} \neq \sigma^2_{i,b}$ 
for $a \neq b$).
Since this geometric update is performed mainly with 
the MSCKF features, it should be performed in the same EKF
update as the reprojection error update --
as the MSCKF features will be marginalized directly afterwards.
Furthermore, any feature track location $[u ~ v]^\top$
used in this update is ignored later on to prevent reusing information.
In practice, the depth measurements across images 
are correlated 
at least to some degree due to sharing the same network weights; however, this can  be
remedied by inflating the noise.

%
Note that the geometric observations of depth map $ {}^{C_j}\hat{\bD} = f(\bM_j, \bS_j, \hat{\bc}_j)$ are related with the sparse depth map $\bS_j$, which is generated by projecting previously marginalized MSCKF features
onto image plane, and thus never includes
${}^{A}\mathbf{p}_{f_i}$. 
Therefore, we consider the image and sparse depth map 
as constant prior information in our system.
Specifically, they only serve to increase the accuracy
of the zero code depth map estimate and subsequently 
the quality of the zero code as initial
guess for $\mathbf{c}_j$, which need to be good in order for the EKF update with FEJ
Jacobians to work properly.

\subsubsection{Depth Consistency Update}

The depth values of the tracked features should be consistent among different keyframes. 
To jointly improve the quality of recovered dense depths,
we formulate a depth consistency measurement for all
tracked feature locations that were not already used in the sparse geometric update.
For a pair of feature tracks observed as 
$\mathbf{o}_{i} = [u_i ~ v_i]^\top, 
\mathbf{o}_{j} = [u_j ~ v_j]^\top$ in the image planes
of camera frames $\{C_a\}$ and $\{C_b\}$,
the depth consistency residual becomes
\begin{align}
	r^{cons}_{i,j,a,b} 
	&= {}^{C_b}D_j
	- \mathbf{e}_3^\top w(\hat{\mathbf{x}}_{ab},{}^{C_a}D_i,\mathbf{o}_i) \notag \\
	&= {}^{C_b}\hat{D}_j
	- \mathbf{e}_3^\top w(\hat{\mathbf{x}}_{ab},
	{}^{C_a}\hat{D}_i+n_i,\mathbf{o}_i) + n_j \notag \\
	&\approx \mathbf{W}_{{c}_b} \tilde{\mathbf{c}}_{b}
	+ \mathbf{W}_{{x}_{ab}} \tilde{\mathbf{x}}_{ab}
	+ \mathbf{W}_{{c}_a} \tilde{\mathbf{c}}_{a} + n_{i,j} 
	\label{eq:depth consistency error}
\end{align}
where ${}^{C_a}D[u_i,v_i]$ is denoted by ${}^{C_a}D_i$ for clarity,
and similarly for $b$ and $j$.
Additionally, $w$ projects the image coordinate $\mathbf{o}_i$
and depth ${}^{C_a}D_i$ into the $\{C_a\}$ frame,
and then into the $\{C_b\}$ frame, and thus is a function of calibration states, as well as the camera clones
corresponding to $\{C_a\}$ and $\{C_b\}$, which are all encapsulated by
in $\mathbf{x}_{ab}$ here.
The Jacobians $\mathbf{W}_{c_b},\mathbf{W}_{x_{ab}},\mathbf{W}_{c_a}$
are, as before, the measurement with respect to each of the subscripted 
states' error states.
Again, the Jacobians for the codes require a single row of the
Jacobian of the VAE decoder with respect to the code.
Similar to the sparse geometric update the variances from the 
noises $n_i, n_j$ are taken from the UNet's estimated uncertainty map,
and the variance for the compounded noise $n_{i,j}$ is calculated
with a simple covariance propagation.

\subsection{Efficient Network Jacobian Computation}
\label{sec:network_Jacobian}

It is important to note that all of the previously mentioned
measurement updates involving the depth codes require the 
Jacobian $\left. \frac{\partial \bD}{\partial \tilde{\mathbf{c}}}       
\right\vert_{\hat{\mathbf{c}}}$ 
of the predicted depth map with respect to the code vector error state
-- evaluated at the current code estimate.
This is definitely a difficult task to accomplish in real
time, as deep learning libraries are optimized to retrieve gradient vectors, but not full Jacobian matrices.
Czarnowski et al. \cite{czarnowski2020deepfactors} 
report that the Jacobian computation
in TensorFow takes over 300ms
on a GTX 1080Ti GPU, which is much too slow for our needs.

Since we only require the Jacobian of the decoder with respect
to the input depth code -- and not all the intermediate weights
-- we actually do not need to calculate a full chain rule. 
Instead, we propose to simply estimate the Jacobian using the following finite difference equation.
\begin{align}
	\label{eq:network jacobian}
	\scalemath{1}{
		\left. \frac{\partial \bD}{\partial \tilde{c}_i} \right\vert_{\hat{c}_i}
		\approx \frac{f(\hat{\bc}
			+ \delta \mathbf{e}_{i})
			- \hat{\bD}}{\delta} 
	}
\end{align}
for some typically small $\delta$, and where $\mathbf{e}_i$ is equivalent to the $i$th column of the $c \times c$ identity matrix $\mathbf{I}$.
This equation represents the derivative of the depthmap for a single element $c_i$ of the code 
$\mathbf{c} \in \mathbb{R}^{c \times 1}$.
Since the depth decoder function $f$ 
is a neural network, which is designed to process batch data,
it seems perfectly suited for this application -- which only requires
running a mini batch of size $c$ ($c=32$ in our case) for just the lightweight VAE decoder, as the UNet skip connections are reused from predicting the initial zero-code depth estimate.
The proposed method can estimate the full dense 
Jacobian with near-perfect
precision (as shown later in the experiments) while only requiring a single batched forward pass through the network.
We found that this operation only takes roughly 10ms
on a desktop-grade GPU (i.e., GTX 1080Ti or RTX 2070 super)
and 40ms on the GTX 1060 on one of our laptops.
This is over an order of magnitude speedup from using
the chain rule Jacobians calculated by a network backward pass.

\section{Experimental Results} \label{sec:exp}
%


\begin{table} [t]
	\caption{
		Performance of depth prediction and encoding methods on the NYU Depth V2 dataset. RMSE and iRMSE are in meters and
		$m^{-1}$, respectively. GPU and CPU timings are in millseconds.
	}
	\label{tab:network}
	\begin{adjustbox}{width=\columnwidth,center}
		\LARGE
		\begin{tabular}{cccccccccc} \toprule
			\textbf{Methods}& \textbf{RMSE} $ \downarrow$
			&  \textbf{iRMSE} $ \downarrow$ & \textbf{MAPE} $\downarrow$ & 
			$\delta_1 \uparrow$ & $\delta_2 \uparrow$ & $\delta_3 \uparrow$ & \textbf{GPU} 
			& \makecell[c]{\textbf{CPU} \\ 4 Threads} 
			& \makecell[c]{\textbf{CPU} \\ 1 Thread} \\\midrule
			FastDepth-RGB&  0.456& 0.080&  0.115& 0.873& 0.968& 0.991& \textbf{2.517} & \textbf{14.313}& \textbf{23.482}\\
			Ours-RGB    &  0.545&  0.090& 0.130& 0.830& 0.951& 0.985& 3.713 & 21.568& 36.407\\
			Ours-RGB-Sp & 0.316& \textbf{0.052}& \textbf{0.066}& \textbf{0.944}& \textbf{0.986}& \textbf{0.996}&3.745 & 21.563& 36.302\\
			Ours-Gray & 0.535& 0.089& 0.133& 0.833& 0.953& 0.986& 3.796& 21.520& 36.218 \\ 
			Ours-Gray-Sp & \textbf{0.315}& 0.055 & 0.071&  0.939& 0.985& 0.995& 3.746 & 21.537& 36.400 \\
			\bottomrule
		\end{tabular}
	\end{adjustbox}
	\vspace*{-2em}
\end{table}


\begin{table} [!b]
	\vspace{-6pt}
	\caption{
		Performance of the depth estimation on EuRoC dataset.
		Entries for ``zero code``/``updated'' prefixed with ``Sp''
		have grayscale and sparse depth input,
		while those without have just grayscale input.
		For each sequence, the method with the best performance is marked in green
		with the second best in blue.
		RMSE and iRMSE are in meters and
		$m^{-1}$, respectively.
	}
	\label{tab:tb_exp_euroc_depth}
	\begin{adjustbox}{width=\columnwidth,center}
		\Large
		\begin{tabular}{crcccccccc} \toprule
			& & \textbf{RMSE} $\downarrow$
			& \textbf{iRMSE} $\downarrow$ & \textbf{MAPE} $\downarrow$ &
			$\delta_1 \uparrow$ & $\delta_2 \uparrow$ & $\delta_3 \uparrow$ & \\
			\midrule
			\multirow{5}{1.5cm}{\centering V1-01}
			& VIO (sparse) & \gr{0.2402} & \gr{0.0499} & \gr{0.0457} & \gr{0.9666} & \gr{0.9791} & \gr{0.9847} \\
			& zero code & 0.8168 & 0.1706 & 0.2464 & 0.5222 & 0.8002 & 0.9295 \\
			& updated & 0.8001 & 0.1692 & 0.2413 & 0.5304 & 0.8073 & 0.9338 \\
			& Sp zero code & 0.4867 & 0.0932 & 0.1108 & 0.8644 & 0.9489 & 0.9771 \\
			& Sp updated & \bl{0.4676} & \bl{0.0912} & \bl{0.1073} & \bl{0.8699} & \bl{0.9521} & \bl{0.9788} \\
			\midrule
			\multirow{5}{1.5cm}{\centering V1-02}
			& VIO (sparse) & \gr{0.3991} & \gr{0.0783} & \gr{0.0923} & \gr{0.8990} & \gr{0.9428} & \gr{0.9688} \\
			& zero code & 0.8037 & 0.1584 & 0.2417 & 0.5122 & 0.8153 & 0.9443 \\
			& updated & 0.7923 & 0.1573 & 0.2386 & 0.5192 & 0.8200 & 0.9460 \\
			& Sp zero code & 0.7212 & 0.1245 & 0.1992 & 0.7599 & 0.8944 & 0.9500 \\
			& Sp updated & \bl{0.6018} & \bl{0.1177} & \bl{0.1697} & \bl{0.7865} & \bl{0.9185} & \bl{0.9643} \\
			\midrule
			\multirow{5}{1.5cm}{\centering V1-03}
			& VIO (sparse) & \gr{0.4631} & \gr{0.0650} & \gr{0.1032} & \gr{0.8831} & \gr{0.9448} & \gr{0.9775} \\
			& zero code & 0.8907 & 0.1722 & 0.2603 & 0.4293 & 0.7505 & 0.9150 \\
			& updated & 0.8830 & 0.1708 & 0.2580 & 0.4334 & 0.7555 & 0.9175 \\
			& Sp zero code & 0.8309 & 0.1108 & 0.2354 & 0.7109 & 0.8710 & 0.9440 \\
			& Sp updated & \bl{0.6871} & \bl{0.1032} & \bl{0.1981} & \bl{0.7393} & \bl{0.9020} & \bl{0.9646} \\
			\midrule
			\multirow{5}{1.5cm}{\centering V2-01}
			& VIO (sparse) & \gr{0.5056} & \gr{0.1125} & \gr{0.1123} & \gr{0.8642} & \gr{0.9293} & \gr{0.9603} \\
			& zero code & 0.9356 & 0.2160 & 0.2828 & 0.4330 & 0.7321 & 0.9061 \\
			& updated & 0.9219 & 0.2161 & 0.2794 & 0.4378 & 0.7368 & 0.9090 \\
			& Sp zero code & 0.6940 & 0.1192 & 0.1726 & 0.7647 & 0.9006 & 0.9550 \\
			& Sp updated & \bl{0.6556} & \bl{0.1167} & \bl{0.1629} & \bl{0.7732} & \bl{0.9084} & \bl{0.9598} \\
			\midrule
			\multirow{5}{1.5cm}{\centering V2-02}
			& VIO (sparse) & \gr{0.5723} & \gr{0.0978} & \gr{0.1175} & \gr{0.8792} & \gr{0.9281} & \gr{0.9545} \\
			& zero code & 1.3989 & 0.2516 & 0.3552 & 0.2899 & 0.5364 & 0.7371 \\
			& updated & 1.3892 & 0.2501 & 0.3525 & 0.2937 & 0.5407 & 0.7403 \\
			& Sp zero code & 0.8595 & 0.1292 & 0.2272 & 0.7064 & 0.8680 & 0.9379 \\
			& Sp updated & \bl{0.7767} & \bl{0.1248} & \bl{0.2055} & \bl{0.7198} & \bl{0.8828} & \bl{0.9485} \\
			\midrule
			\multirow{5}{1.5cm}{\centering V2-03}
			& VIO (sparse) & \gr{0.4876} & \gr{0.0717} & \gr{0.1129} & \gr{0.8778} & \gr{0.9583} & \gr{0.9804} \\
			& zero code & 1.1357 & 0.2353 & 0.3254 & 0.2867 & 0.5705 & 0.8053 \\
			& updated & 1.1306 & 0.2340 & 0.3241 & 0.2890 & 0.5730 & 0.8084 \\
			& Sp zero code & 0.6923 & 0.0993 & 0.1877 & 0.7450 & 0.9158 & 0.9675 \\
			& Sp updated & \bl{0.6516} & \bl{0.0970} & \bl{0.1771} & \bl{0.7557} & \bl{0.9245} & \bl{0.9727} \\
			\bottomrule
		\end{tabular}
	\end{adjustbox}
\end{table}

\begin{table*}[t]
	\caption{
		Pose Estimation Evaluation: RMSE of ATE (deg/meters) on EuRoC.
	}
	\vspace*{-1.0em}
	\label{tab:tb_exp_euroc_pose}
	\begin{center}
		\resizebox{.77\linewidth}{!}{
			\begin{tabular}{cccccccc} \toprule
				\textbf{Methods}& \textbf{V1-01} & \textbf{V1-02} & \textbf{V1-03} & \textbf{V2-01} & \textbf{V2-02} & \textbf{V2-03} & \textbf{Average} \\\midrule
				OpenVINS & 1.138/0.056& 1.860/0.072& 2.138/0.069&  0.908/0.098& 1.190/0.061& 1.356/0.286   & 1.432/0.107\\
				CodeVIO &  1.157/0.054& 1.866/0.071& 2.167/0.068&  0.894/0.097& 1.197/0.061 & 1.241/0.275& \bf{1.420}/\bf{0.104} \\
				\bottomrule
			\end{tabular}
		}
	\end{center}
	\vspace*{-2.5em}
\end{table*}

\subsection{Evaluation of Depth Prediction}
\begin{figure}[tb]
\vspace*{-1em}
\centering
\newcommand{\figWidth}{ .12 \columnwidth} 
\newcommand{\figa}{000170}
\newcommand{\figb}{000538} 
\newcommand{\figc}{001246} 
\setlength\tabcolsep{0.3mm} 
\scriptsize
\begin{tabular}{ c c c c c c c c}
  	\begin{minipage}[c]{\figWidth}\centering
  		\includegraphics[width=\columnwidth]{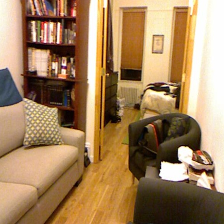}
  	\end{minipage}
  	&
  	  	\begin{minipage}[c]{\figWidth}\centering
  		\includegraphics[width=\columnwidth]{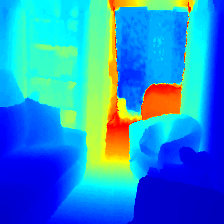}
  	\end{minipage}
  	&
  	  	\begin{minipage}[c]{\figWidth}\centering
  		\includegraphics[width=\columnwidth]{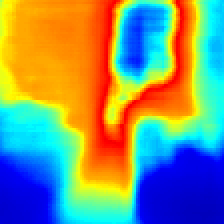}
  	\end{minipage}
  	&
  	  	\begin{minipage}[c]{\figWidth}\centering
  		\includegraphics[width=\columnwidth]{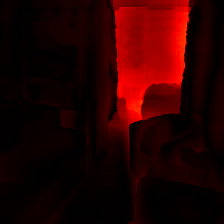}
  	\end{minipage}
  	&
  	  	\begin{minipage}[c]{\figWidth}\centering
  		\includegraphics[width=\columnwidth]{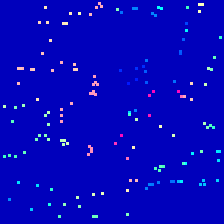}
  	\end{minipage}
  	&
  	  	\begin{minipage}[c]{\figWidth}\centering
  		\includegraphics[width=\columnwidth]{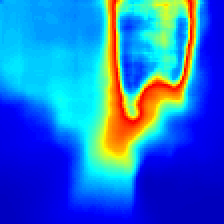}
  	\end{minipage}
  	&
  	  	\begin{minipage}[c]{\figWidth}\centering
  		\includegraphics[width=\columnwidth]{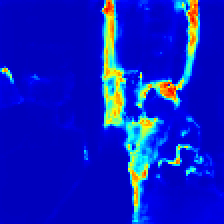}
  	\end{minipage}
  	&
  	  	\begin{minipage}[c]{\figWidth}\centering
  		\includegraphics[width=\columnwidth]{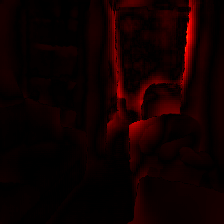}
  	\end{minipage}
\\
\begin{minipage}[c]{\figWidth}\centering
	\includegraphics[width=\columnwidth]{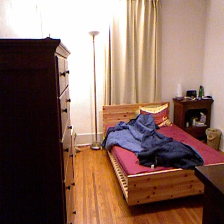}
\end{minipage}
&
\begin{minipage}[c]{\figWidth}\centering
	\includegraphics[width=\columnwidth]{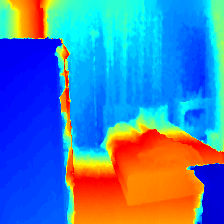}
\end{minipage}
&
\begin{minipage}[c]{\figWidth}\centering
	\includegraphics[width=\columnwidth]{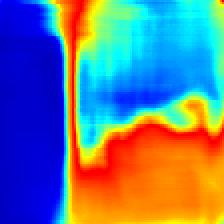}
\end{minipage}
&
\begin{minipage}[c]{\figWidth}\centering
	\includegraphics[width=\columnwidth]{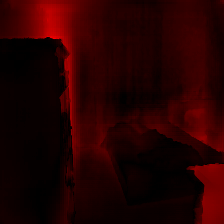}
\end{minipage}
&
\begin{minipage}[c]{\figWidth}\centering
	\includegraphics[width=\columnwidth]{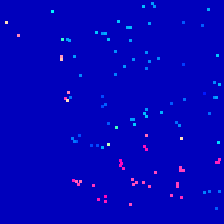}
\end{minipage}
&
\begin{minipage}[c]{\figWidth}\centering
	\includegraphics[width=\columnwidth]{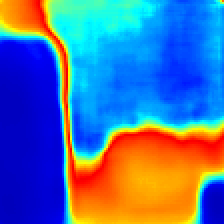}
\end{minipage}
&
\begin{minipage}[c]{\figWidth}\centering
	\includegraphics[width=\columnwidth]{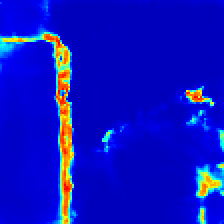}
\end{minipage}
&
\begin{minipage}[c]{\figWidth}\centering
	\includegraphics[width=\columnwidth]{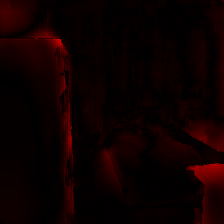}
\end{minipage}
\\
\begin{minipage}[c]{\figWidth}\centering
	\includegraphics[width=\columnwidth]{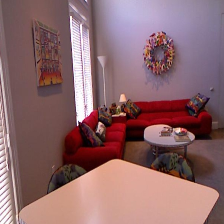}
\end{minipage}
&
\begin{minipage}[c]{\figWidth}\centering
	\includegraphics[width=\columnwidth]{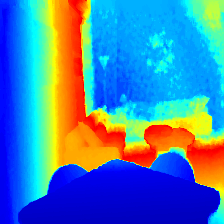}
\end{minipage}
&
\begin{minipage}[c]{\figWidth}\centering
	\includegraphics[width=\columnwidth]{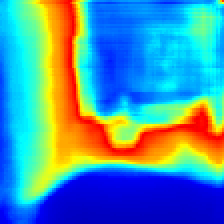}
\end{minipage}
&
\begin{minipage}[c]{\figWidth}\centering
	\includegraphics[width=\columnwidth]{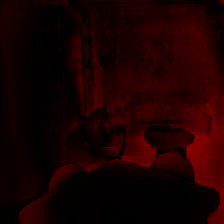}
\end{minipage}
&
\begin{minipage}[c]{\figWidth}\centering
	\includegraphics[width=\columnwidth]{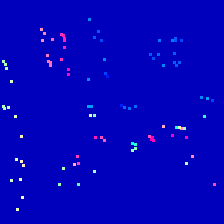}
\end{minipage}
&
\begin{minipage}[c]{\figWidth}\centering
	\includegraphics[width=\columnwidth]{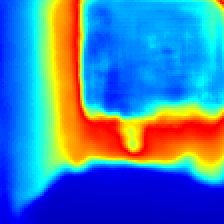}
\end{minipage}
&
\begin{minipage}[c]{\figWidth}\centering
	\includegraphics[width=\columnwidth]{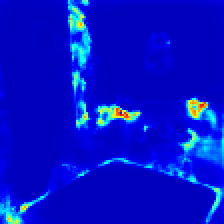}
\end{minipage}
&
\begin{minipage}[c]{\figWidth}\centering
	\includegraphics[width=\columnwidth]{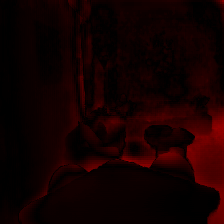}
\end{minipage}
\\
\end{tabular}
\caption{
Performance of neural networks. From left to right side: input RGB image, ground-truth depth, predicted depth of Ours-RGB, RMSE of Ours-RGB depth prediction, sparse depth input of Ours-RGB-sp (125 fast points), predicted depth of Ours-RGB-sp, predicted depth uncertainty of Ours-RGB-sp, RMSE of Ours-RGB-sp depth prediction.
}
\vspace*{-2em}
\label{fig:predctions-nyu}
\end{figure}

We extensively evaluate the network used in this work, regarding the depth prediction accuracy and run time.  All the evaluations in this section are conducted on a commercial desktop with a GTX 1080Ti GPU and Intel i7-8086k CPU@4.0GHz. 
Some standard metrics from the literature are used to evaluate the accuracy of depth prediction, including root mean squared error of depth (RMSE),  root mean squared error of inverse depth (iRMSE),  mean absolute percentage error (MAPE), 
and $\delta_i$, which is the percentage of predicted depth with a certain relative error~\cite{Eigen2014_NIPS}.
To evaluate the computational intensity, the run time evaluated on different hardwares are shown, including the 1080Ti GPU, only CPU with 4 threads, and only CPU with one single thread. The results are shown in Table~\ref{tab:network}. Five networks are compared including FastDepth~\cite{wofk2019fastdepth} with input of RGB image only (\textit{FastDepth-RGB}), the proposed depth-wise convolution based CVAE with RGB input only (\textit{Ours-RGB}), the proposed CVAE with both RGB and sparse depth (\textit{Ours-RGB-Sp}), the proposed CVAE with grayscale image only (\textit{Ours-Gray}), and the proposed CVAE with both grayscale image and sparse depth (\textit{Ours-Gray-Sp}).
The input images are with a resolution of $224 \times 224$.
In the evaluations, the number of sparse depth values of fast points fed into networks is always 125, and noise-free.
Although the proposed CVAE networks without sparse depth perform slightly worse in accuracy and computational intensity than FastDepth, the advantage is that the dense depth (size $224^2$) can be encoded into a compact and optimizable representation of only size $32$. 
%
The results also suggest that the scale determined sparse depth can improve the image only networks in a large margin with a negligible increment of computation.
Representative depth predictions from the network are also shown in Fig.~\ref{fig:predctions-nyu}. It is clear that sparse depth aided prediction has significantly smaller errors. The predicted uncertainty fit the error of depth prediction well, which indicates the Unet stream in the proposed CVAE can predict the depth uncertainties reasonably.
%

\subsection{Accuracy of the Estimated Network Jacobian}
\begin{figure}
	\centering
	\includegraphics[width=.8\columnwidth]{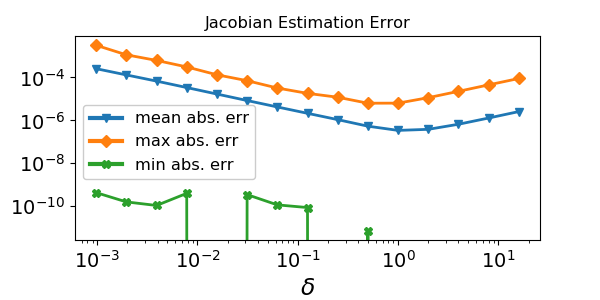}
	\caption{The accuracy of our finite difference Jacobian compared to 
		those from PyTorch autograd by the absolute difference.}
	\label{fig:fd_jac}
	\vspace*{-1.4em}
\end{figure}

Typically, in finite difference equations Eq.~\eqref{eq:network jacobian}, a very small $\delta$ must
be used to perturb the inputs in order to obtain reasonable accuracy
(not to be confused with the $\delta_i$ accuracy metric in the other
tables).
However, in our case, we found that a small $\delta$ actually made the 
Jacobian less accurate than the chain rule Jacobian taken from PyTorch
autograd.
In order to find the best $\delta$ value, we use PyTorch
autograd in batch mode to calculate the first 32 rows of the Jacobian using the batch
input we used for the finite difference Jacobian (clearly
before applying the perturbation).
We only do the first 32 rows so that we can reuse the same batch
as our finite difference method, as calculating the full
Jacobian with PyTorch would require many more forward passes.
We test with the first 100 images in the EuRoC V2\_01 sequence, and
compare the max, min, and average element-wise absolute error between the 
autograd Jacobian and the first 32 rows of ours over a range of
different $\delta$ values.
Fig.~\ref{fig:fd_jac} summarizes our findings, where it can be seen that
the accuracy of the finite difference Jacobian is best with $\delta$
between roughly 0.5 and 1.0, which is many orders of magnitude higher than
we initially expected from observing other implementations of finite
difference Jacobians.%
\footnote{ Note that we based our initial implementation off of
	the \texttt{fdjac} example code provided by \cite{Driscoll_FNC},
	which uses a $\delta$ value on the order of $10^{-8}$.
}
We suspect that this is due to the fact that a small perturbation in the 
input code does not change the output by much, since the CVAE decoder
maps a 32-dimensional vector to $224^2$ outputs, so it must compress
information about large groups of pixels into each single code entry.
Due to this, we chose $\delta=0.666$ in our experiments, as it falls 
in the most accurate range of $\delta$ values. 
\subsection{Full Evaluation on EuRoC Dataset}

\begin{figure}
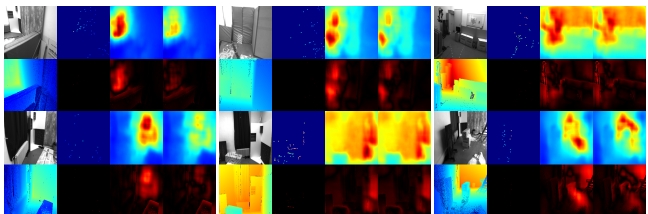
 
	\newcommand{\w}{.325\columnwidth}
	\centering
	\foreach \seq/\stamp in {V1_01/1403715340862142976,
		V1_02/1403715564912143104,
		V1_03/1403715907884058112} {
		\begin{subfigure}[t]{\w}
			
			\includegraphics[width=\columnwidth]{%
				figures/euroc/\seq/comparison_GOOD_\stamp} 
		\end{subfigure} \hspace{-9pt}
	}

	\centering
	\foreach \seq/\stamp in {V2_01/1413393244605760512,
		V2_02/1413393920105760512,
		V2_03/1413394939755760384} {
		\begin{subfigure}[t]{\w}
			
			\includegraphics[width=\columnwidth]{%
				figures/euroc/\seq/comparison_GOOD_\stamp} 
		\end{subfigure} \hspace{-9pt}
	}
	
	\caption{
		Some qualitative results from the EuRoC dataset.
		Each example is formatted in columns of image/ground truth,
		and then three columns of input sparse depth, zero code prediction, and 
		updated depth with the error map below each depth map.
		The top is for each of the V1 sequences, 
		and the V2 is on the bottom.
	}
	\vspace{-2em}
	\label{fig:euroc_pretty_pics}
\end{figure}

In order to gauge the performance of the whole system, we test it on 
all the Vicon Room sequences (see Fig.~\ref{fig:overview}) of the EuRoC dataset~\cite{burri2016euroc}.
These sequences include an accurate LiDAR point cloud map, which
we used to obtain the ground truth depths.
While some works have focused on the reconstruction accuracy
compared to the whole LiDAR point cloud map on this dataset \cite{Rosinol2020_ICRA},
we instead are only concerned with the accuracy of the depth maps
-- as we do not perform any global mapping.
To this end, we used the ground truth camera poses to project the 
point cloud into the undistorted image plane -- keeping 
the minimum depth per location in order to account for occlusions.
Note that in these experiments, we only used the network trained on the NYUv2
dataset (``\textit{Ours-Gray}'' and ``\textit{Ours-Gray-Sp}'' from Table \ref{tab:network})  in order to demonstrate 
the potential generalization of our method.

A quantitative analysis of the proposed depth estimation system can be observed in
Table \ref{tab:tb_exp_euroc_depth}. 
We evaluate the sparse VIO depth estimates as well as our grayscale networks trained on the 
NYUv2 dataset as previously mentioned, both before and after the code updates.
In general, the sparse VIO depths, while noisy, are more accurate than any network predictions.
This is expected, since we need accurate sparse depths to improve the network accuracy
over image-only input, and we would not expect the depth completion network to be more
accurate than the input.
Second to the VIO depths across the board, the proposed dense depth estimation 
with grayscale and sparse depth input after code updates is the most accurate
of the network predictions.
While the proposed sparse geometric and
depth consistency updates generally improve the depth maps over the 
zero code depths, the inclusion of sparse depths makes even the zero
code prediction more accurate than the updated image-only depth map.
Additionally, the relative improvement between zero code and updated for the 
sparse depth network is typically better than that of the image-only network.
This shows that the sparse depth inputs improve
the value of the zero code as the initial estimate, and thus
the FEJ Jacobian is more reasonable to use.
A visualization of the input and predictions for the sparse depth network can be seen in Fig.~\ref{fig:euroc_pretty_pics} as well as in Fig.~\ref{fig:overview}.

Another interesting fact is that the accuracy of pose estimation (Table~\ref{tab:tb_exp_euroc_pose}) is also slightly improved with the learned optimizable dense depth. This demonstrates the pose estimation and dense depth mapping can potentially promote each other. Note that, other than the dense
depth code estimation, CodeVIO and OpenVINS are using the same
estimator parameters here.

\begin{figure}
	\includegraphics[width=\columnwidth]{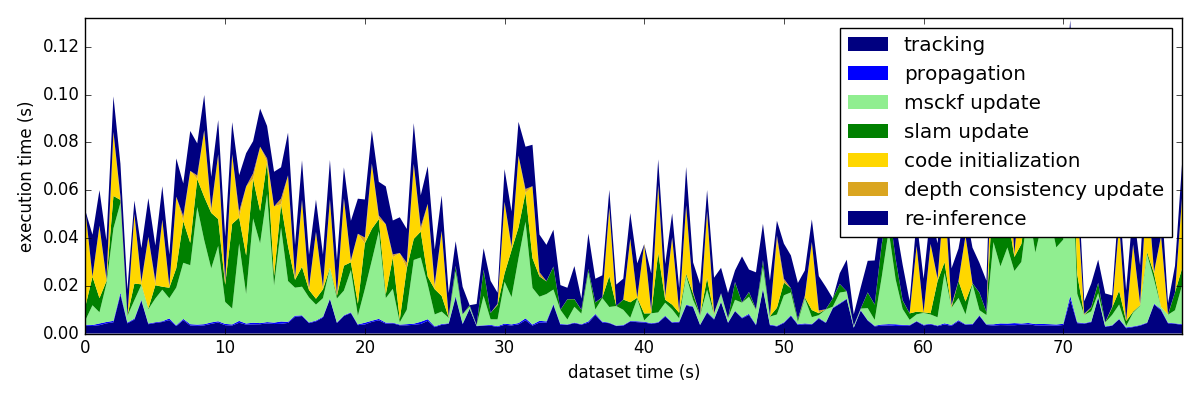}
	\caption{The runtime of CodeVIO on EurocMav V1-03 sequence.}
	\vspace*{-2em}
	\label{fig:timing}
\end{figure}

The run time of the main processes in the proposed VIO system is shown in Fig.~\ref{fig:timing}. The main stages include sparse feature tracking, IMU propagation, temporary MSCKF feature update with sparse geometry, ``long-lived"  SLAM feature update, zero code initialization (network inference and Jacobian calculation), depth consistency update, and the reinference of dense depth maps after their codes are updated. The mean run time for the entire loop is $0.0447\pm 0.0226$ seconds, which is suitable for applications requiring real-time capability. It should be noted that our current implementation is not highly-optimized and can still be further improved. 
%

\section{Conclusions and Future Work} \label{sec:conc}

In this paper, we have designed a real-time VIO system with the capability of dense local mapping.
An efficient CVAE neural network for depth prediction and encoding is proposed, which also takes advantage of the sparse depth from VIO and generalize well on a completely unseen dataset.
Dense depth is compactly encoded in a low dimensional code, which is optimizable and estimated with inertial navigation states in a tightly-coupled lightweight MSCKF framework.
In addition, a novel and efficient network Jacobian computation method is proposed, which is over an order of magnitude speedup over the chainrule method.
%
This work targets local dense mapping, and we leave full SLAM with a global dense depth map as future work.

{
\def\bibfont{\scriptsize}
\printbibliography
%
}

\end{document}